\documentclass[10pt,twocolumn,letterpaper]{article}
\pdfoutput=1
\usepackage{iccv}
\usepackage{amsmath}
\usepackage{amssymb}
\usepackage{times}
\usepackage{epsfig}
\usepackage{graphicx}
\usepackage{multirow}

\usepackage[breaklinks=true,bookmarks=false]{hyperref}

\iccvfinalcopy 

\def\iccvPaperID{2206} 

\ificcvfinal\fi
\begin{document}

\title{Grouped Spatial-Temporal Aggregation for Efficient Action Recognition }

\author{Chenxu Luo \qquad \qquad Alan Yuille\\
Department of Computer Science, The Johns Hopkins University, Baltimore, MD 21218, USA\\
{\tt\small \{chenxuluo,ayuille1\}@jhu.edu}
}

\maketitle

\begin{abstract}
Temporal reasoning is an important aspect of video analysis. 3D CNN shows good performance by exploring spatial-temporal features jointly in an unconstrained way, but it also increases the computational cost a lot. Previous works try to reduce the complexity by decoupling the spatial and temporal filters. In this paper, we propose a novel decomposition method that decomposes the feature channels into spatial and temporal groups in parallel. This decomposition can make two groups focus on static and dynamic cues separately. We call this grouped spatial-temporal aggregation (GST). This decomposition is more parameter-efficient and enables us to quantitatively analyze the contributions of spatial and temporal features in different layers. We verify our model on several action recognition tasks that require temporal reasoning and show its effectiveness. 
\end{abstract}

\section{Introduction}
With the success of convolutional neural networks in image classification~\cite{simonyan2014two,he2016deep}, action recognition has also shifted from traditional hand-crafted features (e.g. IDT~\cite{wang2013action}) to deep learning based methods. With the introduction of large scale datasets \cite{sigurdsson2016hollywood,carreira2017quo,goyal2017something} and more powerful models \cite{carreira2017quo,wang2018non}, deep network based methods have become standard for video classification tasks.

Temporal reasoning plays an important role in video analysis. However, common video datasets used for action recognition, such as UCF101~\cite{soomro2012ucf101} and Kinetics~\cite{carreira2017quo}, do not require much temporal reasoning. Most of the classes in the datasets can be recognized based only on static scenes or objects~\cite{li2018resound}. 
Furthermore, some works even show that shuffling the temporal ordering, the accuracy remains almost the same~\cite{xie2018rethinking}. This suggests that models trained on those datasets may not necessarily exploit temporal cues.

\begin{figure}[!t]
	\begin{center}
		\includegraphics[width=0.9\linewidth]{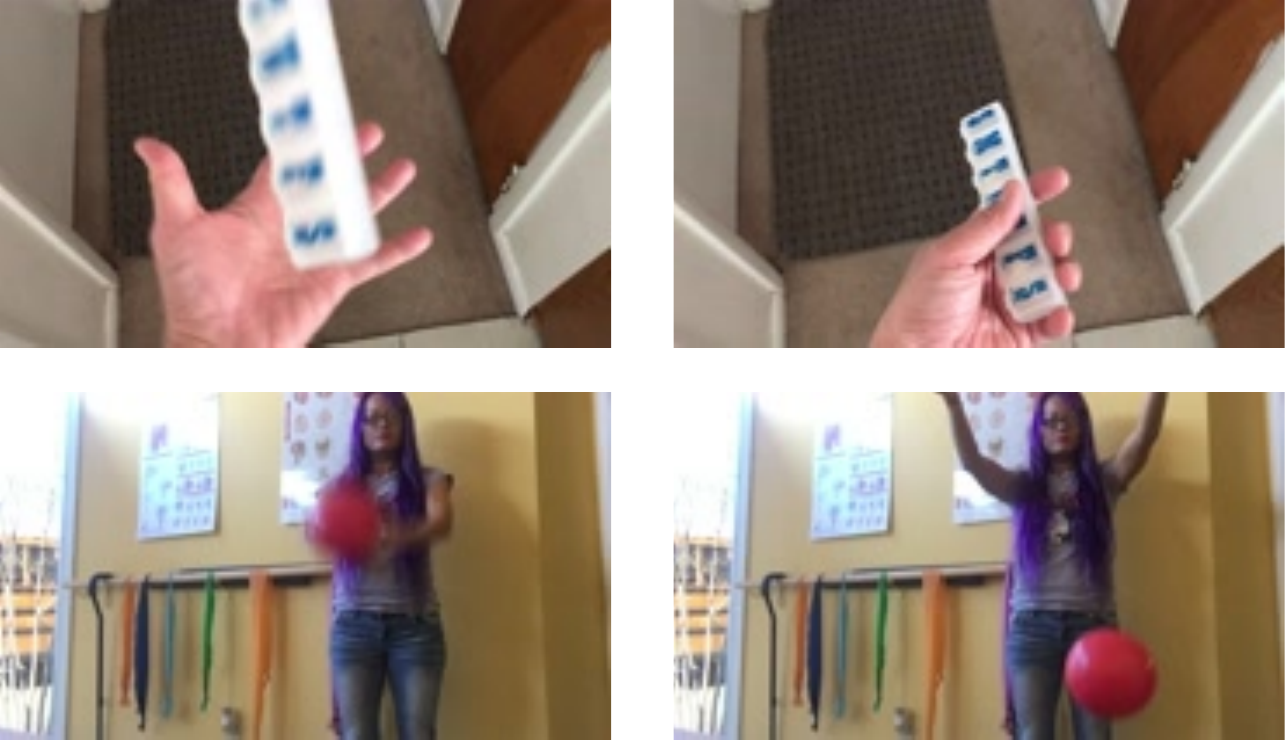}
	\end{center}
	\caption{Examples from the something-something dataset~\cite{goyal2017something}. The groundtruth for the two videos are ``throwing something in the air and catching it'' and ``throwing something in the air and letting it fall''. This requires temporal information to correctly differentiate these two classes. 
	}
	\label{fig:dataset_example}
\end{figure}

Recently, several datasets~\cite{goyal2017something,damen2018scaling,li2018resound} have been released which require temporal modeling. For example, Figure \ref{fig:dataset_example} shows two examples from the Something-Something dataset~\cite{goyal2017something}. Seeing only a single frame is not sufficient to determine the class. The two examples are similar at the beginning (the first column) but have different results at the end (see the second column). These datasets emphasizes on the temporal aspects in action recognition. 

However, this does not mean that the static information in each frame is not helpful. Appearance encodes rich cues for temporal reasoning. For example, in Figure \ref{fig:dataset_example}, we can narrow down the possible interpretations by only seeing a single frame. And we can infer the action from the sparsely sampled frames by observing the state changes.

Existing spatial-temporal networks, such as C3D~\cite{tran2015learning} and I3D~\cite{carreira2017quo}, learn spatial and temporal features jointly in an unconstrained way. Although they can achieve good performance, they also introduce a large number of parameters that results in computational burdens. Some works~\cite{sun2015human,qiu2017learning,Tran_2018_CVPR,xie2018rethinking} try to reduce the cost by decomposing a 3D convolutional kernel into spatial and temporal part separately. However, it remains unsure how spatial and temporal information is utilized in a network. 

In this paper, we propose to decompose along the channel dimension instead and show that it is more parameter-efficient than previous methods. Our method is inspired by the widely used group convolutions. 
The intuition here is that some channels may be more related to spatial features and some channels focus more on motion features, by analogy to the different functions of neurons (\eg Parvo and Magno cells) in the retina. In previous methods, the spatial and temporal features are entangled together across channels. And directly applying the same operator on all channels may not be optimal and efficient. So we propose to decompose the feature maps into a spatial group and a channel group and apply different operations respectively. Based on this, we design a two-path module in each residual block. Different from previous works where the groups are symmetric, we use one path to model spatial information and the other path to explore temporal information. After that, the spatial-temporal features are concatenated. We call this Grouped Spatial-Temporal aggregation (GST). Unlike the cascaded decomposition used in the P3D-like networks~\cite{qiu2017learning}, our method implements it in a parallel way, which can exploit features in a more efficient way. This spatial-temporal decomposition not only reduces the parameters but also facilitates the network to learn different aspects (\ie static and dynamic information) and temporal multi-scale features separately in a single layer. 

 

Unlike previous works that model spatial-temporal information in an unconstrained way, our decomposition allows us to analyze how networks exploit spatial and temporal features in different layers. Interestingly, we find that low level features focus more on static cues while high level features focus more on dynamic cues when the networks are trained on temporal modeling tasks. The networks can automatically learn a soft selection without any further constraints. 

The proposed module can be easily inserted into any common 2D networks such as ResNet~\cite{he2016deep}. We conduct experiments on several datasets that require temporal information. Our model can outperform existing methods with less computational cost.

To summarize, our contributions include (a) We propose a novel decomposition method for 3D convolutional kennels that explicitly model spatial and temporal information separately and efficiently; (b) We quantitatively analyze the contribution of spatial and temporal features in different layers; (c) We achieve the state-of-the-art results on several datasets that require temporal modeling with much less computational costs. 

\section{Related Works}
\paragraph{Datasets for Action Recognition}
The prevalent datasets such as UCF101~\cite{soomro2012ucf101} or Kinetics~\cite{carreira2017quo} have strong static bias and focus less on temporal orders~\cite{li2018resound,xie2018rethinking,zhou2018temporal}. Li~\etal\cite{li2018resound} quantitative evaluate the bias towards static representations, such as scenes and objects. Such biases distract researcher from exploring better temporal model. It remains unsure whether the model trained on these datasets actually learn the action itself or simply exploit the bias.

Recently, crowd-acted and fine-grained  datasets~\cite{goyal2017something,sigurdsson2016hollywood,damen2018scaling,fouhey2018lifestyle} receive more and more favor and attention. These newly collected datasets pose new challenges for action recognition. 
Especially fine-grained video datasets such as Something-Something~\cite{goyal2017something,mahdisoltani2018fine}, Jester, Diving48~\cite{li2018resound} require extensive temporal modeling. For example, the two classes ``tearing something into two pieces" and ``tearing something just a little bit'' in something-something~\cite{goyal2017something} can not be determined without seeing the whole sequence. 
\begin{figure}[!t]
	\begin{center}
		\includegraphics[width=0.95\linewidth]{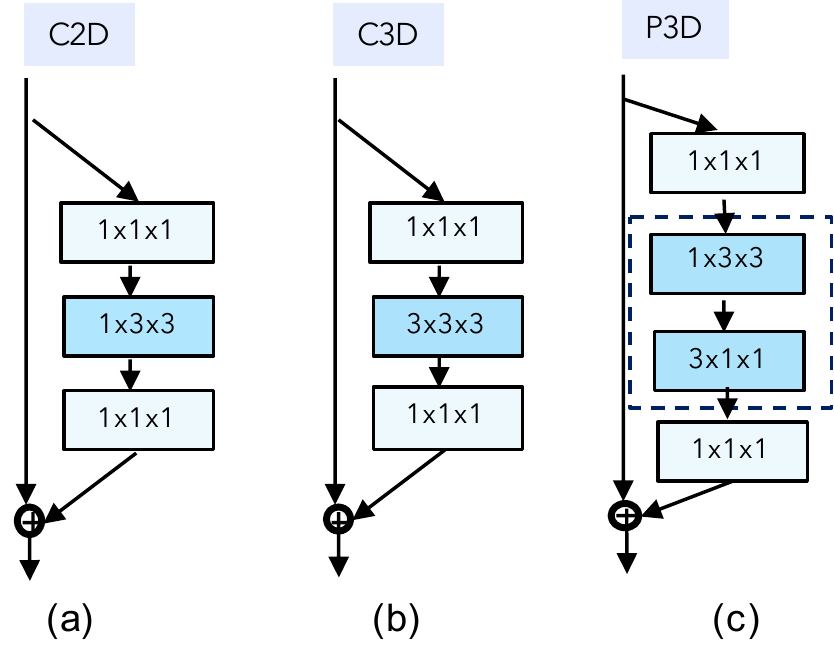}
	\end{center}
	\caption{Comparison between three common types of networks. (a) shows a 2D network, TSN~\cite{wang2016temporal} and TRN~\cite{zhou2018temporal} belongs to this category. (b) shows an C3D type network. (c) shows a P3D block (also known as S3D or R(2+1)D), which decouple the spatial and temporal filters.}
	\label{fig:p3d}
\end{figure}
\begin{figure*}[!t]
	\begin{center}
		\includegraphics[width=\linewidth]{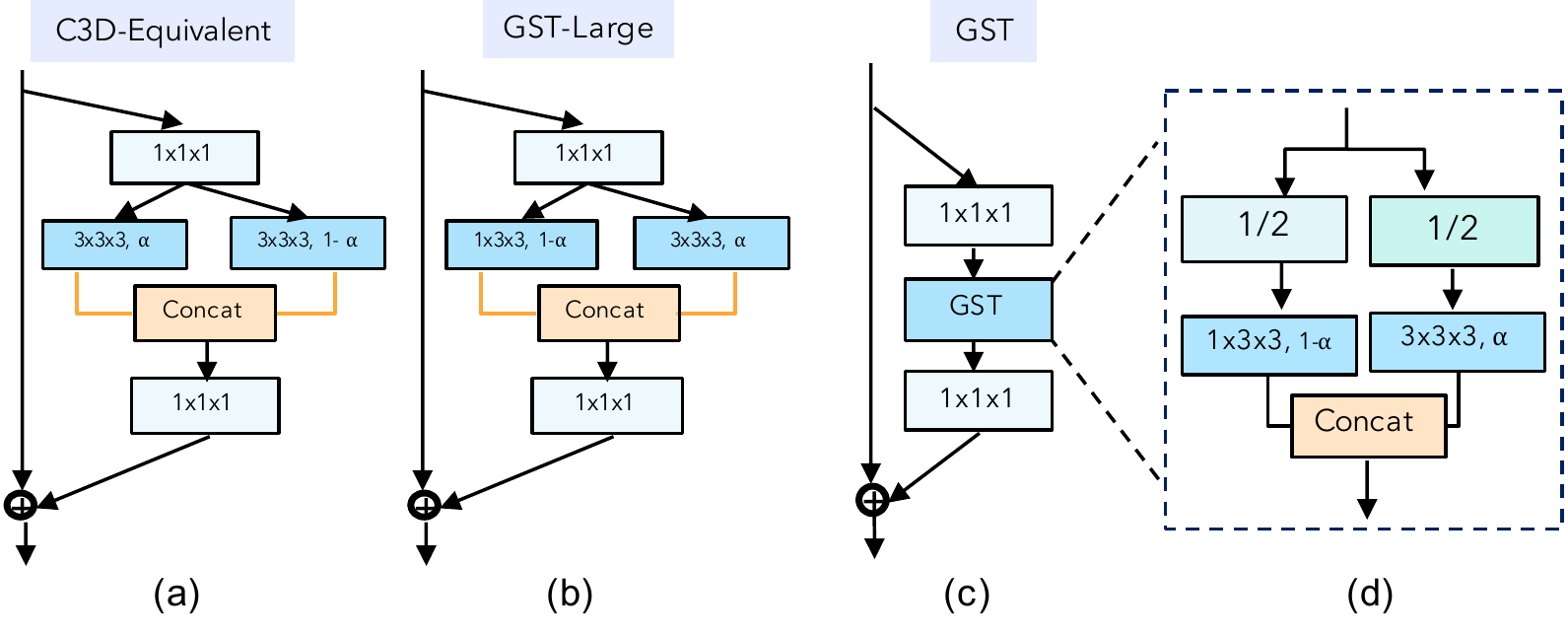}
	\end{center}
	\caption{Overview of our proposed method. (a) shows an equivalent network of C3D. (b) shows that replacing one path to spatial only convolutions, denotes as GST-Large. (c) shows our method and (d) illustrates the proposed GST module. In our GST module, the input feature map is divided into two groups; One group for spatial modeling and the other group for temporal modeling. The two paths use the same number of parameters and are concatenated together. }
	\label{fig:model}
\end{figure*}
\paragraph{Temporal Modeling}
With the success of deep neural networks in visual recognition, a lot of works have been done to extend it for video classification. 
The early works simply apply 2D convolutions on single frames and fuse then. Karpathy~\etal\cite{Karpathy_2014_CVPR} propose several fusion strategies for frame aggregation. Later, TSN~\cite{wang2016temporal} propose a new sampling strategy and use late fusion strategy to aggregate features of each frame. TRN~\cite{zhou2018temporal} improves this by introducing multiscale MLP for temporal aggregation. Both of them use late a fusion strategy. Although these 2D networks perform well on datasets like UCF101~\cite{soomro2012ucf101} or Kinetics~\cite{carreira2017quo}, they show much less satisfactory results on datasets that require extensive temporal reasoning~\cite{goyal2017something,li2018resound}.

In another branch, 3D networks(e.g. C3D~\cite{tran2015learning},  I3D~\cite{carreira2017quo}, P3D~\cite{qiu2017learning}) recently have gained attention. The first 3D network (\ie C3D~\cite{tran2015learning}) has a huge number of parameters and is hard to train. I3D~\cite{carreira2017quo} propose to inflate an ImageNet pre-trained model to 3D by weight copying. Res3D~\cite{hara2018can} systematically evaluates several common inflated structures.  ECO~\cite{zolfaghari2018eco} adds a 3D-ResNet after the 2D network for temporal fusion. 
SlowFast~\cite{feichtenhofer2018slowfast} uses two different architectures operating on different temporal frequencies. Our work explores static and motion features in the channel dimension.

\paragraph{Optical Flow for Action Recognition} Starting from the seminal work of Two-stream network~\cite{simonyan2014two}, optical flow has been widely used for motion representation. 
Most works find their model can perform better when combining optical flow in addition to RGB as input. However, computing optical flow can be time-consuming and is independent of the network. Some works try to jointly optimize optical flow estimation and the  classification network~\cite{zhu2017hidden,fan2018end}, or implicitly model optical flow in RGB network~\cite{Sun_2018_CVPR}.
\paragraph{Efficient Temporal Modeling}
Standard 3D networks like C3D~\cite{Tran_2018_CVPR} contains a huge number of parameters that are difficult to train. Sun~\etal~\cite{sun2015human} reduce the parameters by decoupling spatial and temporal kernels. P3D~\cite{qiu2017learning} and S3D~\cite{xie2017aggregated} further explore it with different architectures. R(2+1)D~\cite{Tran_2018_CVPR} shows that this can achieve better results with the same number of parameters as 3D Convolutions. Figure \ref{fig:p3d} shows the comparison between these common structures. 

TSM~\cite{lin2018temporal} replace temporal filters with shift modules. This simple way does not introduce new parameters and can perform surprisingly well on temporal modeling tasks. 

\section{Approach}

An overview of our proposed method is shown in Fig.\ref{fig:model}, the output channels are split into two groups, one for spatial modeling, and the other for spatial-temporal modeling. The spatial part is just the standard 2D convolutions. For the temporal part, 3D convolutions are used. Then the spatial-temporal features are concatenated together. In this way, we can use even fewer parameters than a standard 2D network counterpart (such as a ResNet-50~\cite{he2016deep}) but can significantly boost its ability for temporal modeling. 
In the following sections, we describe our novel Grouped Spatial-Temporal aggregation (GST) module in detail.

\subsection{Decomposing a 3D Convolution Kernel}
Consider a 3D convolutional kernel with $C_i$ input channels and $C_o$ output channels. $T,H,W$ are the kernel size along the temporal and spatial dimensions respectively. The kernel is of size  $C_{o}\times C_{i}\times T\times H\times W$, which is $T$ times larger than its 2D counterpart. Given that modern CNNs such as ResNet~\cite{he2016deep} usually have a large number of channels, this significantly increases the cost. 

There are a lot of works that seek to reduce the parameters by factorizing the convolutional kernels. One widely used way is to decouple the spatial and temporal part~\cite{sun2015human,qiu2017learning,Tran_2018_CVPR,xie2018rethinking}. The underlying assumption here is that the spatial and temporal kernels are orthogonal to each other. Mathematically, we can write this decomposition as 
\begin{equation}
    w= w_t\times w_s
\end{equation}
where $w_s\in\mathbb{R}^{C_{o}\times C_{i}\times 1\times H\times W}$ and $w_t\in\mathbb{R}^{C_{o}\times C_{i}\times T\times 1\times 1}$ are the spatial and temporal kernels respectively. 
R(2+1)D~\cite{Tran_2018_CVPR} shows that this decomposition can achieve better performance under the same number of parameters used as 3D convolutions. 

\subsection{Grouped Spatial-temporal Decomposition}
Group convolution has been widely used in image recognition, for example, ResNext~\cite{xie2017aggregated}, ShuffleNet~\cite{zhang2018shufflenet}, to name a few. However, in video tasks, it has been less explored. Most of the existing works simply replace the original convolutions with group convolutions, such as Res3D~\cite{hara2018can}.

 However, as shown in the experiments, directly applying group convolutions in a trivial way, which results in symmetric groups, cannot bring improvements. So we propose to decompose the large 3D convolutional filter along the channel dimension in an asymmetric way.

Since both appearance and motion are useful for action recognition,  some feature channels may focus more on static appearance while other channels may focus more on dynamic motion features. So modeling them separately is effective and efficient. Based on this assumption, we propose to let the two groups of features model spatial and temporal information separately. 

Figure \ref{fig:model} (a) shows an equivalent network architecture to C3D (Figure \ref{fig:p3d}(b)), where the output channels are split into two groups and then concatenated. In our GST design, we apply spatial-only convolutions (\ie 2D convs) to the first group of features and spatial-temporal convolutions (\ie 3D convs) to the other group. We denote this as GST-Large as shown in Figure \ref{fig:model}(b). To further reduce the number of parameters, we decompose the input channels into two groups, spatial and temporal, and apply 2D and 3D convolutions respectively. This can encourage the channels in each group to concentrate on static semantic features and dynamic motion features separately and thus easier for training. Static and dynamic features can be thus combined in a natural way. Formally, our decomposition module GST can be written as
\begin{equation}
    w_{_{GST}}=w_{gs}\oplus w_{gt}
\end{equation}
where $w_{gs}\in\mathbb{R}^{C_{o_s}\times C_{i_s}\times 1\times H\times W}$ is used for the spatial path and $w_{gt}\in\mathbb{R}^{C_{o_t}\times C_{i_t}\times T\times H\times W}$ is used for the temporal path. Here, $o_s$ and $i_s$ are the number of output and input channels for spatial path, and $o_t$ and $i_t$ for the temporal path in the same way. Our method enables multi-scale temporal modeling in a single layer. 
In the experiments, we show that this spatial-temporal decomposition enjoys better parameter utilization and can effectively reduce the number of parameters while leading to better performance.

\subsection{Computational Costs for the Spatial and Temporal Path}\label{sec:number_channels}
To control the complexity of the GST, we introduce two parameters to specify the complexity of spatial and temporal branches. We use $\alpha$ to specify the proportion of temporal output channels and $\beta$ to specify the number of input channels for spatial and temporal features.

For output channels, we have $C_{o_t}=\alpha C_{o}$ number of channels for the temporal path, and the rest for the spatial path, so the total number of parameters of the spatial and temporal path are: $(1-\alpha)HWC_{i_s}C_{o}$ and $\alpha THWC_{i_t}C_{o}$ respectively. Larger values of $\alpha$ results in more channels for temporal modeling and thus higher computation cost. While smaller $\alpha$ means lower capacity for temporal path and thus lower complexity. As pointed out in SlowFast~\cite{feichtenhofer2018slowfast}, lower channel capacity means weaker ability to represent spatial semantics. We carry out experiments with $\alpha=1/2,1/4,1/8$ respectively. Empirically, we find that fewer temporal channels are beneficial for reducing the computation cost without hurting the performance. In section \ref{sec: spatial-vs-temporal}, we quantitatively analyze how spatial and temporal channels are utilized in each block. 

For input channels, if we set $\beta=1$, then $C_{i_s}=C_{i_t}=C_{i}$ and both spatial and temporal path take as input the whole feature maps. We denote this model as GST-Large (Figure \ref{fig:model} (b)). Compared with the model in Figure \ref{fig:model} (a), which is equivalent to 3D convolutions, we replace one path for spatial modeling. This allows multi-scale temporal modeling in a single layer. In the experiments, we show that this not only reduces the parameters, but also improves the performance.

For more efficient architectures, we set $\beta=1/2$, so $C_{i_s}=C_{i_t}=C_{i}/2$. The models are shown in Figure \ref{fig:model} (c) and (d), where the input channels are split evenly into two groups and one group is used for spatial modeling and the other group is used for temporal modeling. 
With the commonly used kernel size $H=W=T=3$, our GST models have roughly the same or even less number of parameters than a 2D network with properly designed spatial-temporal channel decomposition. However, our model contains sufficient temporal interactions and thus has higher temporal modeling ability than merely using a 2D network. 

To summarize, we list the number of parameters for different architectures in Table \ref{tab:parameters}.
\begin{table}[!h]
    \centering
    \begin{tabular}{|c|l|}
    \hline
     Model    &  \# params \\ \hline
     C2D    & $H\cdot W\cdot C_{i}\cdot C_{o}$  \\
     C3D    & $T\cdot H\cdot W\cdot C_{i}\cdot C_{o}$  \\
     P3D    & $(H\cdot W+T)\cdot C_{i}\cdot C_{o}$  \\
     C3D(groups=g)  &  $T\cdot H\cdot W/g\cdot C_{i}\cdot C_{o}$  \\
     GST-Large & $ (1-\alpha + \alpha T)HWC_{i}C_{o}$ \\
     GST  & $(1-\alpha + \alpha T)HWC_{i}C_{o}/2$  \\ \hline
    \end{tabular}
    \caption{Comparison of the number of parameters for each spatial-temporal block. }
    \label{tab:parameters}
\end{table}

\subsection{Network Architecture}
The proposed GST module is flexible and can be easily plugged into most of the current networks. More specifically, we replace each of the $3\times3$ convolutional layer with our GST module while keeping other layers unchanged. The final prediction is a simple average pooling of each frame. We show that this can already achieve good results since the spatial-temporal features are frequently aggregated in each intermediate block. This is contrary to the late fusion method TRN~\cite{zhou2018temporal}, which needs a complex fusion module that operates on the high-level features.

\section{Experiments}
\subsection{Datasets}
We evaluate our method on five video datasets that require temporal modeling.\\
\textbf{Something-Something} Something v1~\cite{goyal2017something} and v2~\cite{mahdisoltani2018fine} are two large scale video datasets for action recognition. There are totally about 110k(v1) and 220k(v2) videos for 174 fine-grained classes with diverse objects, backgrounds, and viewpoints. The fine-grained level classes need extensive temporal reasoning to differentiate them as shown in the example in Fig \ref{fig:dataset_example}. We mainly conduct experiments and justify each component on these two datasets.\\
\textbf{Diving48} Diving48~\cite{li2018resound} is a newly released dataset with more than 18K video clips for 48 diving classes. This requires more focus on pose and motion dynamics. In fact, this dataset aims to minimize the bias towards static frames and facilitate the study of dynamics in action recognition. We report the accuracy on the official train/val split. \\
\textbf{Egocentric Video Datasets} We also evaluate our model on two egocentric video tasks to show that our proposed model is generic on a variety of tasks. We use two recently collected egocentric dataset, Epic Kitchen~\cite{damen2018scaling} and EGTEA Gaze+~\cite{Li_2018_ECCV}. For Epic Kitchen, we report the verb classification results using the same split as ~\cite{Baradel_2018_ECCV}. EGTEA Gaze++ is a recently collected dataset with approximately 10K samples of 106 activity classes. We use the first split as~\cite{Li_2018_ECCV}, which contains 8299 training and 2022 testing instances. 

\subsection{Implementation Detail}
We implement our model in Pytorch. We adopt ResNet-50~\cite{he2016deep} pretrained on Imagenet~\cite{russakovsky2015imagenet} as the backbone. The parameters of temporal paths are randomly initialized. 

For the temporal dimension, we use the sparse sampling method described in TSN~\cite{wang2016temporal}. And for spatial dimension, the short side of the input frames are resized to 256 and then cropped to $224\times 224$. We do random cropping and flipping as data augmentation during training time.

We train the network with a batch-size of 24 on 2 GPUs and optimize using SGD with an initial learning rate of 0.01 for about 40 epochs and decay it by a factor of 10 every 10 epochs. The total training epochs are about 60. The dropout ratio is set to be 0.3 as in ~\cite{wang2018videos}.

During the inference time, we sample the middle frame in each segment and do center crop for each frame. We report the results of \textbf{single crop} unless specified. 

\subsection{Results on Something-Something Datasets}
We first evaluate each component of our model on both something-something v1 and v2 datasets. 
\paragraph{Ablation Study} We conduct several ablation studies on the Something-Something V1 and V2 validation sets~\cite{goyal2017something}.  For all the models, we sample 8 frames using the same sampling method as TSN~\cite{wang2016temporal} and use ResNet-50~\cite{he2016deep} as backbone network. Results are shown in Table \ref{tab:v1-ablation}.

We compare our model with three baselines, ResNet50 based C3D, C3D with group convolutions and P3D. For C3D and P3D, we use the architecture depicted in Fig. \ref{fig:p3d} (b) and (c) respectively, and for C3D with groups of 2, we set each $3\times3\times3$ convolution to be a group convolution with  group size of 2.  We also compare networks with different spatial and temporal channel ratios ($\alpha=1/2,1/4,1/8$ described in Sec.\ref{sec:number_channels}). 
\begin{table}[!h]
\centering
\begin{tabular}{ccccccc}
\hline
\multirow{2}{*}{Method} & \multirow{2}{*}{\#params} & \multicolumn{2}{c}{v1} &  & \multicolumn{2}{c}{v2} \\ \cline{3-4} \cline{6-7} 
 &           & top1       & top5      &  & top1       & top5      \\ \hline
\multicolumn{1}{l}{C3D$_{3\times3\times3}$} & 42.5M &46.2   &   75.6 & &  60.9 & 87.0  \\ \hline
\multicolumn{1}{l}{C3D \small{groups=2}} & 29.6M &45.1 & 74.0   & & 59.9 & 86.5   \\ 
\multicolumn{1}{l}{P3D} & 29.4M & 45.7 & 75.0 & & 59.8 & 85.8 \\
\multicolumn{1}{l}{GST\small{-Large(1/4)}} & 29.6M & \textbf{47.7} & \textbf{76.4} & & \textbf{62.0}  & \textbf{87.5} \\ \hline 
\multicolumn{1}{l}{C2D} & 23.9M & 20.4 & 48.1 &  & 30.5 & 61.2  \\ 
\multicolumn{1}{l}{GST ($\alpha$=1/2)}  & 23.9M&  46.7    &  \textbf{76.2} & &  61.4 & \textbf{87.3}   \\
\multicolumn{1}{l}{GST ($\alpha$=1/4)}   & \underline{21.0M} &   \textbf{47.0}    &   76.1 & & \textbf{61.6} & 87.2   \\
\multicolumn{1}{l}{GST ($\alpha$=1/8)}  & \textbf{19.7M}  &   46.7    &   75.7 & &  60.7 & 86.6   \\ \hline
\end{tabular}
\caption{ Ablation Study on Something v1 and v2 validation set. For all the models, we use a ResNet-50 based backbone and sample 8 frames for each video clip.}
\label{tab:v1-ablation}
\end{table}

\begin{table*}[!t]
\tabcolsep=0.3cm
\centering
\begin{tabular}{lllllc}
\hline
Model       & Backbone                                    & \#Frame & GFLOPs &  Top1 & Top5                  \\ \hline
TRN-2stream~\cite{zhou2018temporal} & BN-Inception                                & 8      &    -   & 42.0 & - \\ \hline
\multirow{2}{*}{ECO~\cite{zolfaghari2018eco}}         & \multirow{2}{*}{BNInception+
3D ResNet-18} & 8      & 32    & 39.6 &     -                  \\
       &                                             & 16     & 64      & 41.4 &  -                     \\ \hline
MFNet-C50~\cite{lee2018motion} & ResNet50 & 10 & - & 40.3 & 70.9 \\ 
MFNet-C101~\cite{lee2018motion}  & ResNet101 & 10 & - & 43.9 & 73.1 \\ \hline
NL I3D~\cite{wang2018non} & 3D ResNet-50 & 32$\times$2 clips & 168$\times$2  & 44..4 & 76.0 \\ 
NL I3D+GCN~\cite{wang2018videos}  & 3D ResNet-50                                & 32$\times$2 clips     &   -             & 46.1 & 76.8                  \\ \hline
TSM~\cite{lin2018temporal}         & ResNet-50                                   & 8      &   33       & 43.4 & 73.2                  \\
TSM~\cite{lin2018temporal}          & ResNet-50                                   & 16     &  65       & 44.8 & 74.5       \\   \hline        
S3D~\cite{xie2018rethinking} & BN-Inception & 64 &  66.38  & 47.3 & 78.1\\
S3D-G~\cite{xie2018rethinking} & BN-Inception & 64 & 71.38  & 48.2 & \textbf{78.7} \\ \hline
GST (ours)        & ResNet-50                                   & 8      &  \textbf{29.5}      & 47.0 & 76.1                  \\
GST (ours) & ResNet-50 & 8$\times$2 clips & 29.5$\times$2 & 47.6 & 76.6 \\ 
GST (ours)        & ResNet-50                                   & 16     &  59         & \textbf{48.6} & 77.9  \\  \hline
\end{tabular}
\caption{Comparison with state-of-the-art results on the Something V1 validation set. We mainly consider the methods that only take RGB as input for fair comparison. For each model, we report its top 1 and top 5 accuracy as well as its FLOPs. 
}
\label{tab:sth-v1}
\end{table*}
\begin{table*}[!h]
\centering
\begin{tabular}{llllllll}
\hline
\multicolumn{1}{c}{\multirow{2}{*}{Method}} & \multirow{2}{*}{Frames} & \multicolumn{1}{c}{\multirow{2}{*}{Backbone}} & \multicolumn{1}{c}{Val} &       &  & \multicolumn{2}{c}{Test} \\ \cline{4-5} \cline{7-8} 
\multicolumn{1}{c}{}                        &                         & \multicolumn{1}{c}{}                          & Top-1                   & Top-5 &  & Top-1       & Top-5      \\ \hline
TRN~\cite{zhou2018temporal}                                         & 8                       & BN-Inception                                  & 48.8                    & 77.6  &  & 50.9        & 79.3       \\
TSM~\cite{lin2018temporal}                                         & 8                       & Resnet-50                                     & 59.1*                   &85.6* &  &    \multicolumn{1}{c}{-}        &       \multicolumn{1}{c}{-}     \\
TSM~\cite{lin2018temporal}                                         & 16                      & Resnet-50                                     & 59.4*                   & 86.1* &  & 60.4*       & 87.3*      \\ \hline
GST (ours)                                        & 8                       & Resnet-50                                     & 61.6                    & 87.2  &  & 60.04*      &{87.17}*     \\
GST (ours)                                        & 16                      & Resnet-50                                     & \textbf{62.6}                    & \textbf{87.9}  &  &   \textbf{61.18}*           &      \textbf{87.78}*       \\ \hline
TRN-2stream~\cite{zhou2018temporal}                                 & 8                       & BN-Inception                                  & 55.5                    & 83.1  &  & 56.2        & 83.2       \\
TSM-2stream~\cite{lin2018temporal}                                 & 16                      & Resnet-50                                     & 63.5                    & 88.6  &  & 64.3       & 90.1    \\ \hline   
\end{tabular}
\caption{Comparison with state-of-the-art results on the something-something v2 dataset. * denotes results of 5 crops}
\label{tab:somethingv2}
\end{table*}
First, for our GST-Large model, we set $\alpha=1/4$. This results in a similar number of parameters as P3D or naive 3D group convolutions with a group size of 2. However, our model outperforms other methods on both datasets. Even compared with the larger C3D model, it still performs much better. This shows that our parallel decomposition can better utilize the parameters than the cascaded way like P3D.  Also, compared with the original 3D convolutions, GST-Large uses only partial channels for temporal modeling and thus reduces the computational costs significantly. However, our model generalizes better than C3D by decomposing the channel space into spatial and temporal separately. 

Second, for more efficient models, our proposed GST uses a similar amount of parameters as a 2D ResNet-50, but performs much better than 2D models. This shows that our model allocates the parameter space more efficiently. Compared with 3D group convolutions, we show that replacing one of the groups with spatial-only convolutions is beneficial. Even compared with the C3D networks, our model combining spatial and temporal cues still performs better on both v1 and v2 dataset. This shows that a 3D network contains redundancy and empirically, we find that by separating spatial and temporal channels, the networks are easier to train and generalize better. 

We also study the impact of temporal channel capacity. We experiment with different temporal channel ratios ($\alpha=1/2,1/4,1/8$).  We find that dropping the ratio of temporal channels does not hurt the performance significantly. This shows that maybe lower channel capacity is needed for temporal modeling. In section \ref{sec: spatial-vs-temporal}, we examine in detail how the temporal channel capacities affect spatial-temporal modeling. 

In later experiments, we set $\alpha=1/4$ and $\beta=1/2$ as default, for its good trade-off between accuracy and efficiency. 
\paragraph{Comparison with state-of-the-arts}
The results on v1 and v2 are shown in Table \ref{tab:sth-v1} and Table \ref{tab:somethingv2} respectively. 

On the v1 dataset, our model sampling only 8 frames can already outperform most current methods. Our method outperforms the late fusion method TRN~\cite{zhou2018temporal} and ECO~\cite{zolfaghari2018eco} because it can better encode the spatial and temporal features. Our model can perform as well as S3D using significantly fewer frames and even outperform complex models like non-local network~\cite{wang2018non} with graph convolution~\cite{wang2018videos}.

Compared with v1, v2 is two times larger with fewer label ambiguities. We test it on both validation and test set. Our model again achieves state-of-the-art results. Especially, our single-stream model outperforms two-stream TRN~\cite{zhou2018temporal} by 5\% absolutely. Even though our model takes only RGB as input, our 16-frame model provides competitive results compared with two-stream networks.

\subsection{Results on Diving48 Dataset}
We test our model on Diving48~\cite{li2018resound} dataset. This dataset requires modeling the subtle body motions in order to classify correctly, while the background and object cues seem almost useless. We sample 16 frames from each video clip.
\begin{table}[!h]
\small
\tabcolsep=0.08cm
  \begin{tabular}{lcc}
\hline  
Method & Pre-training & Accuracy \\ \hline
C3D(64 frames)(from~\cite{Li_2018_ECCV}) & - & 27.6 \\
R(2+1)D(from ~\cite{bertasius2018learning}) &  Kinetics & 28.9 \\
R(2+1)D+DIMOFS~\cite{bertasius2018learning} & Kinetics + PoseTrack & 31.4 \\  \hline
C3D-ResNet18(our impl.) & ImageNet & 33.0 \\ 
P3D-ResNet18(our impl.) & ImageNet & 30.8 \\
GST-ResNet18(ours) & ImageNet & \textbf{34.2}    \\ \hline
C3D-ResNet50(our impl.) & ImageNet & 34.5 \\
P3D-ResNet50(our impl.) & ImageNet & 32.4\\
GST-ResNet50(ours) & ImageNet & \textbf{38.8}  \\ \hline
\end{tabular}
\caption{Results on the Diving48 Dataset~\cite{Li_2018_ECCV}}
\label{tab:diving}
\end{table}

In Table \ref{tab:diving}, we present quantitative results on this dataset.  Compared with previous works, our method outperforms all other counterparts, like R(2+1)D network, by a large margin. Especially, by only employing a lightweight backbone, ResNet-18, our model can already outperform the previous state-of-the-art. This shows that our model can efficiently capture important temporal cues. We believe leveraging pose estimation can benefit recognizing diving actions, but this is beyond the scope of this paper. Despite this, our generic model can already outperform current methods.

\subsection{Results on Ego-motion Action Recognition}
To show that our proposed model is generic for various action recognition tasks, we also test it on two recently released ego-motion video datasets,~\ie Epic-Kitchen~\cite{damen2018scaling} and EGTEA Gaze++~\cite{Li_2018_ECCV}. Both datasets focus on activities in the kitchen. So there is less bias towards scenes. For the Epic Kitchen Dataset~\cite{damen2018scaling}, there are a total of 125 verb classes and each verb can be acted on different objects. We report the results on the validation set using the same split as ~\cite{Baradel_2018_ECCV}. We only evaluate on the verb class prediction following ~\cite{Baradel_2018_ECCV} since the main purpose of this paper is on temporal action recognition instead of objects. For the EGTEA Gaze++ dataset, it contains 106 classes with 19 different verbs. We report the results using the split-1 as in ~\cite{Li_2018_ECCV}.

We use the same setting as the experiments on Something-Something datasets and sample 8 frames for each clip and the results are listed in Table. \ref{tab:epic} and \ref{tab:egtea} respectively.

For the Epic-Kitchen dataset, all models use ResNet-50 as the backbone. Our model again achieves better results. 

On EGTEA Gaze++ dataset, we also try a shallow network ResNet-34 as the backbone, for a fair comparison with prior works. Without bells and whistles, our model can even perform better than previous two-stream models with the same backbone architecture. This shows that our proposed module is generic for temporal modeling.

\subsection{Analysis of Spatial and Temporal Features} \label{sec: spatial-vs-temporal}

\begin{figure*}[!t]
	\begin{center}
		\includegraphics[width=\linewidth]{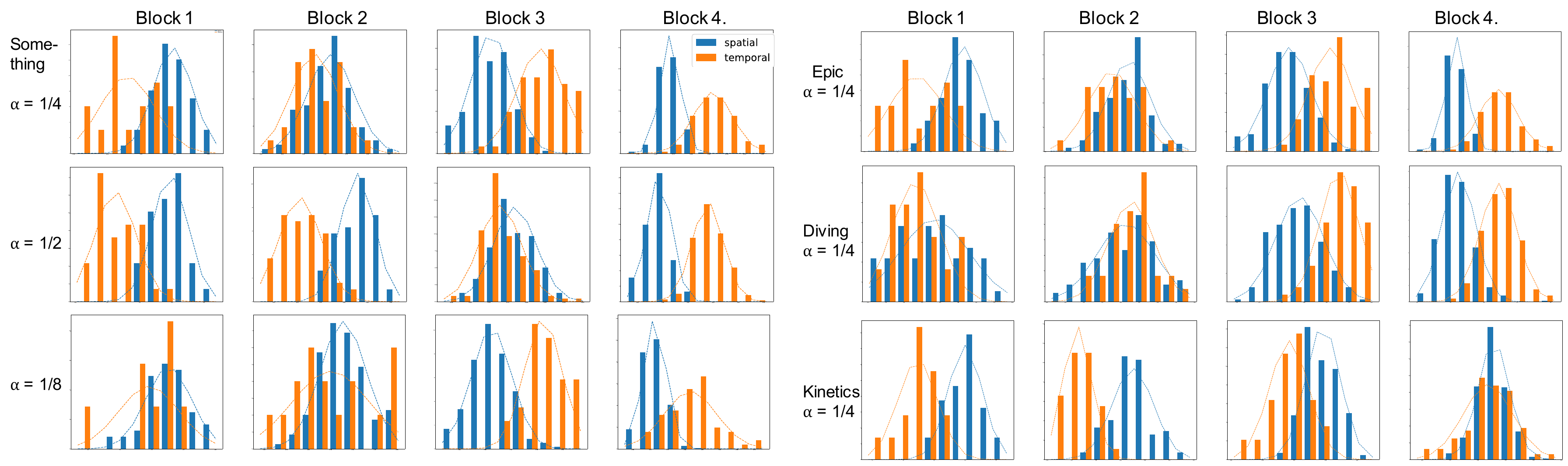}
	\end{center}
	\caption{Contributions of spatial and temporal information. We plot the histograms of weights in each BN layer that corresponds to spatial and temporal group respectively after each GST module. Higher weight means the information in that channel is more important.}
	\label{fig:temporal_weight}
\end{figure*}

To understand how spatial and temporal information is encoded in each layer, we carefully check the weight of the BN layer after each GST module. The input to the BN layer is a concatenation of spatial and temporal feature maps and the scaling factor of each channel in the BN layer can be used to approximately estimate the importance of that channel. For each bottleneck block, we compute the histograms of the scaling factors of each channel that corresponds to spatial or temporal channel and show them in Figure \ref{fig:temporal_weight}. 

\begin{table}[!h]
\centering
\begin{tabular}{|l|l|l|l|}
\hline
Method   &   \cite{Baradel_2018_ECCV}      & LFB~\cite{wu2018long} &    GST(Ours)         \\ \hline
Top1 (Top 5) & 40.89 (-) &  54.4 (81.8)  & \textbf{56.50} (\textbf{82.72}) \\ \hline
\end{tabular}
\caption{Results on validation set of Epic-Kitchen verb classification tasks using the same split as in ~\cite{Baradel_2018_ECCV}}
\label{tab:epic}
\end{table}  
\begin{table}[!h]
\centering
\begin{tabular}{|l|l|}
\hline
Method       & Video Acc \\ \hline
\cite{Li_2018_ECCV}(I3D-2stream)             &       53.3    \\
\cite{sudhakaran2018attention}(R34-2stream) & 62.2      \\ \hline
P3D-R34(our impl.) & 58.1 \\
GST-R34(ours)    & 62.2      \\  \hline
P3D-R50(our impl.) & 61.1 \\
GST-R50(ours)     & \textbf{64.4}      \\ \hline
\end{tabular}
\caption{Results on EGTEA Gaze++ using split 1}
\label{tab:egtea}
\end{table}

The statistics of the scaling factors in the BN layers show that the two groups of channels encode inherently different cues. The network can learn static and dynamic features separately in a single layer and implicitly learn a soft-weighted dynamic channel selection in each block. 

First, in the left column, we show models with different temporal channel ratios $\alpha$ trained on Something-Something. For $\alpha=1/2$, in block 3, the spatial and temporal weights are less distinguishable, showing that too many temporal channels may encode extra static information. This somehow explains why reducing the number of temporal channels can improve accuracy. While for $\alpha=1/8$, it may not have enough capacity for temporal modeling. 

We also visualize the statistic of models trained on Epic-Kitchen, Diving48 and Kinetics. For datasets that require temporal information, we can see that in low-level features, spatial information is more important and in high-level features, temporal information outweighs spatial information. This may due to that object cues in a single frame are often not enough for determining the action. And the temporal channels thus encode abstract motion features other than static features that help recognize actions. However, for Kinetics, spatial and temporal features are less distinguishable. This suggests that the learned temporal features may contain some static features. 

\begin{figure*}[!h]
	\begin{center}
		\includegraphics[width=0.9\linewidth]{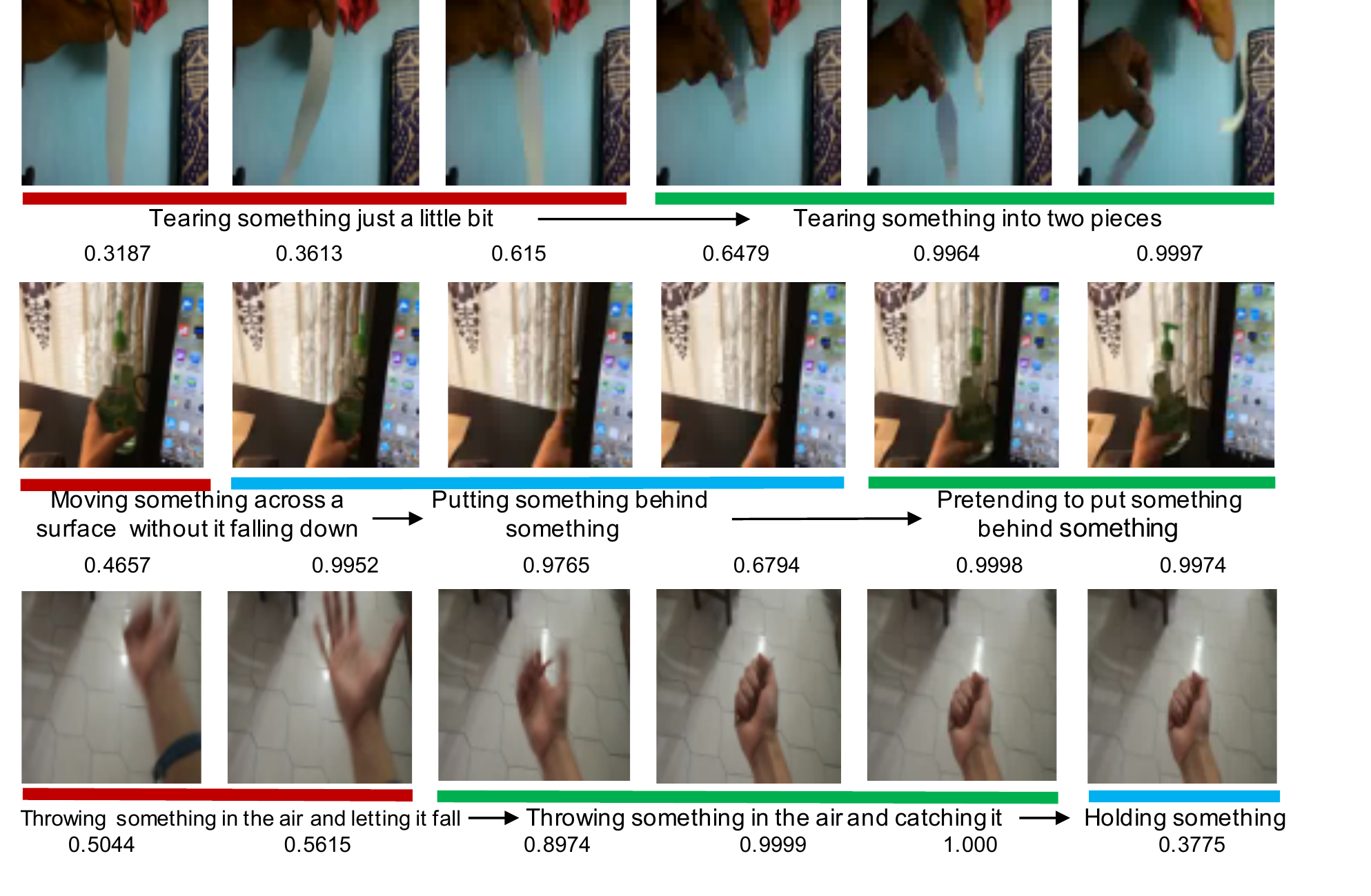}
	\end{center}
	\caption{Examples show how the predictions evolve temporally. We use the 16-frame model trained on Something v2 dataset. We only show six typical frames in each video clip. We compute the prediction of each frame before the average pooling and show the predicted label and confidence score for each frame. Green bars show the correct prediction for the whole video clip. Interestingly, state changes can be discovered without strong supervision.
	}
	\label{fig:pred_example1}
\end{figure*}


Thus, by decoupling spatial and temporal feature channels, we can quantitatively evaluate the contribution of each part. This gives insight into how spatial and temporal cues are encoded from low-level to high-level features, which may benefit future network designs.  

We illustrate some examples from Something v2 val set in Figure \ref{fig:pred_example1}. In each example, we show the network prediction in each intermediate time stamp. Specifically, the final prediction is an average of each frame's prediction. We examine the output in each intermediate frame. Interestingly, the state transitions can be learned given only video-level labels. In the first example, the prediction goes from ``tearing something just a little bit'' to ``tearing something into two pieces'', which corresponds to the state changes of the whole action. Similarly, the network can change to ``pretending to something behind something" after seeing the bottle is moved back. This suggests the state changes of static frames may be crucial to recognize the full action.

\section{Conclusions}
In this paper, we propose a simple yet efficient network for temporal modeling. The proposed GST module decomposes the feature channels into static and dynamic part, and apply spatial and temporal convolutions separately. This decomposition can effectively decrease the computation cost and facilitate the network to explore spatial and temporal features in parallel. Further diagnoses give insight into how the two components contribute to the whole network. 

{\small\textbf{Acknowledgement} This work is partially supported by the Intelligence Advanced Research Projects Activity (IARPA) via Department of Interior/ Interior Business Center (DOI/IBC) contract number D17PC00345.}
{\small
\bibliographystyle{ieee}
\bibliography{egbib}
}

\end{document}